\documentclass[runningheads]{llncs}
\usepackage{graphicx}

\usepackage{tikz}
\usepackage{comment} 
\usepackage{amsmath,amssymb} 
\usepackage{color}

\usepackage[pagebackref=true,breaklinks=true,letterpaper=true,colorlinks,bookmarks=false,urlcolor=red]{hyperref}
\usepackage{gensymb,graphics,subfigure,inputenc}
\usepackage[linesnumbered,algo2e,boxed]{algorithm2e}
\usepackage[figuresright]{rotating}
\usepackage{textcomp,subfigure,multirow,upgreek}
\usepackage{booktabs}
\usepackage{mdwlist}

\usepackage{ragged2e}
\usepackage{fontawesome}
\graphicspath{{Figures/}}

\usepackage{caption}
\begin{document}
\pagestyle{headings}
\mainmatter
\def\ECCVSubNumber{100}  

\title{XingGAN for Person Image Generation} 

\titlerunning{XingGAN for Person Image Generation}
%
\author{Hao Tang\inst{1,2} \and
Song Bai\inst{2} \and
Li Zhang\inst{2} \and 
Philip H.S. Torr\inst{2} \and
Nicu Sebe\inst{1,3}
}
\authorrunning{H. Tang et al.}
%
\institute{$^1$University of Trento (\email{hao.tang@unitn.it}) \\
$^2$University of Oxford \quad
$^3$Huawei Research Ireland}
\maketitle

\begin{abstract}

We propose a novel Generative Adversarial Network (XingGAN or CrossingGAN) for person image generation tasks,~\emph{i.e.},~translating the pose of a given person to a desired one. The proposed Xing generator consists of two generation branches that model the person's appearance and shape information, respectively. Moreover, we propose two novel blocks to effectively transfer and update the person's shape and appearance embeddings in a crossing way to mutually improve each other, which has not been considered by any other existing GAN-based image generation work. Extensive experiments on two challenging datasets,~\emph{i.e.},~Market-1501 and DeepFashion, demonstrate that the proposed XingGAN advances the state-of-the-art performance both in terms of objective quantitative scores and subjective visual realness.
The source code and trained models are available at
\url{https://github.com/Ha0Tang/XingGAN}.

\end{abstract}
\section{Introduction}
\begin{figure}[t]
	\centering
	\includegraphics[width=1\linewidth]{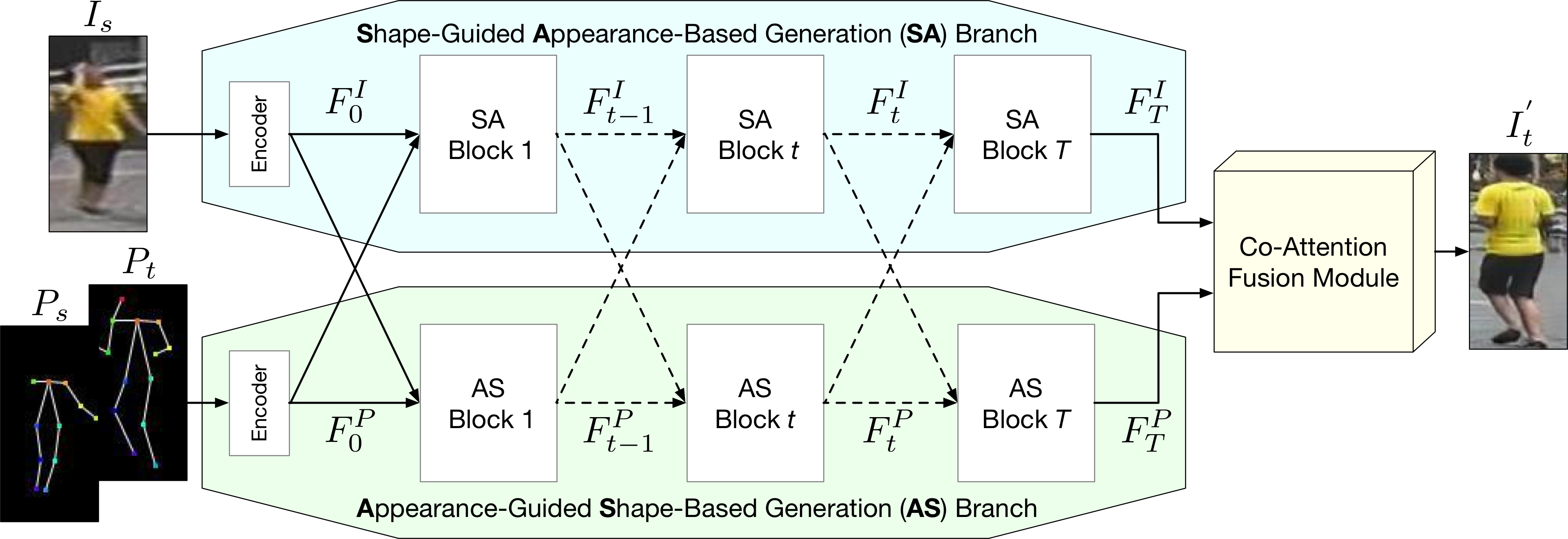}
	\caption{Overview of the proposed Xing generator. Both the Shape-guided Appearance-based generation (SA) and the  Appearance-guided Shape-based generation (AS) branches consist of a sequence of SA and AS blocks in a crossing way. All these components are trained in an end-to-end fashion so that the SA branch and AS branch can benefit from each other to generate more shape-consistent and appearance-consistent person images.}
	\label{fig:method}
\end{figure}

The problem of person image generation aims to generate photo-realistic person images conditioned on an input person image and several desired poses.
This task has a wide range of applications such as person image/video generation \cite{yang2018pose,grigorev2019coordinate,balakrishnan2018synthesizing,ilyes2018pose,liu2019liquid} and person re-identification \cite{zhu2019progressive,qian2018pose}.
Exiting methods such as \cite{ma2017pose,ma2018disentangled,siarohin2018deformable,zhu2019progressive,tang2019cycle} have achieved promising performance on this challenging task.
For example, Zhu \emph{et al.}~\cite{zhu2019progressive} recently proposed a conditional GAN model that comprises a sequence of pose-attentional transfer blocks. Wherein, each block transfers certain regions it attends to and progressively generates the desired person image.

Although \cite{zhu2019progressive} performed an interesting exploration, we still observe unsatisfactory aspects and visual artifacts in the generated person images due to several reasons.
First, \cite{zhu2019progressive} stacks several convolution layers to generate the attention maps of the shape features, then the generated attention maps are used to attentively highlight the appearance features.
Since convolutional operations are building blocks that process one local neighborhood at a time, this means that they cannot capture the joint influence between the appearance and the shape features.
Second, the attention maps in \cite{zhu2019progressive} are only produced by using one single modality, \emph{i.e.}, the pose, leading to insufficiently accurate correlations for both modalities (\emph{i.e.}, the pose and the image modality), and thus misguiding the image generation. 

Based on these observations, we propose a novel Generative Adversarial Network (XingGAN or CrossingGAN), which consists of a Xing generator, a shape-guided discriminator, and an appearance-guided discriminator.
The overall framework is shown in Fig.~\ref{fig:method}. The Xing generator consists of three parts, \emph{i.e.}, a Shape-guided Appearance-based generation (SA) branch, an Appearance-guided Shape-based generation (AS) branch, and a co-attention fusion module.
Specifically, the proposed SA branch contains a sequence of SA blocks, which aim to progressively update the appearance representation under the guidance of the shape representation, while the proposed AS branch contains a sequence of AS blocks, which aim to progressively update the shape representation under the guidance of the appearance representation. 
We also present a novel crossing operation in both SA and AS blocks to capture the joint influence between the image modality and the pose modality by creating attention maps jointly produced by both modalities. 
Moreover, we introduce a co-attention fusion model to better fuse the final appearance and shape features to generate the desired person images.
We present an appearance-guided discriminator and a shape-guided discriminator to jointly judge how likely is that the generated image contains the same person in the input image and how well the generated image aligns with the targeted pose, respectively.
The proposed XingGAN is trained in an end-to-end fashion so that the generation branches can enjoy the mutually improved benefits from each other.

We conduct extensive experiments on two challenging datasets, \emph{i.e.}, Market-1501 \cite{zheng2015scalable} and DeepFashion \cite{liu2016deepfashion}.
Qualitative and quantitative results demonstrate that XingGAN achieves better results than state-of-the-art methods, regarding both visual fidelity and alignment with targeted person poses. 

To summarize, the contributions of our paper are three-fold:
\begin{itemize}
	\item We propose a novel XingGAN (or CrossingGAN) for person image generation. It explores cascaded guidance with two different generation branches, and aims at progressively producing a more detailed synthesis from both person shape and appearance embeddings.
	\item We propose SA and AS blocks, which effectively transfer and update person shape and appearance features in a crossing way to mutually improve each other, and are able to significantly boost the quality of the final outputs.
	\item Extensive experiments clearly demonstrate the effectiveness of XingGAN, and show new state-of-the-art results on two challenging datasets, \emph{i.e.}, Market-1501 \cite{zheng2015scalable} and DeepFashion \cite{liu2016deepfashion}.
\end{itemize}
\section{Related Work}
\noindent \textbf{Generative Adversarial Networks (GANs)} \cite{goodfellow2014generative} consist of a generator and a discriminator where the goal of the generator is to produce photo-realistic images so that the discriminator cannot tell the generated images apart from real images.
GANs have shown the capability of generating photo-realistic images \cite{brock2018large,karras2018style,shaham2019singan}.
However, it is still hard for vanilla GANs to generate images in a controlled setting.
To fix this limitation, Conditional GANs (CGANs) \cite{mirza2014conditional} have been proposed.

\noindent \textbf{Image-to-Image Translation} aims to learning the translation mapping between target and input images. CGANs have achieved decent results in pixel-wise aligned image-to-image translation tasks \cite{isola2017image,tang2018gesturegan,albahar2019guided}.
For example, Isola \emph{et al.} propose Pix2pix~\cite{isola2017image}, which adopts CGANs to generate the target domain images based on the input domain images, such as photo-to-map, sketch-to-image, and night-to-day. 
However, pixel-wise alignment is not suitable for person image generation tasks due to the shape deformation between the input person image and target person image.

\noindent \textbf{Person Image Generation.} To remedy this, several works started to use poses to guide person image generation \cite{ma2017pose,ma2018disentangled,siarohin2018deformable,esser2018variational,tang2019cycle,zhu2019progressive}.
For example, Ma \emph{et al.} first present PG2 \cite{ma2017pose}, which is a two-stage model to generate the target person images based on an input image and the target poses.
Moreover, Siarohin \emph{et al.} propose PoseGAN \cite{siarohin2018deformable}, which requires an extensive affine transformation computation to deal with the input-output misalignment caused by pose differences.
Zhu~\emph{et al.} propose Pose-Transfer \cite{zhu2019progressive}, which contains a sequence of pose-attentional transfer blocks to generate the target person image progressively. 
Besides the aforementioned supervised methods, several works focus on solving this task in an unsupervised setting~\cite{pumarola2018unsupervised,song2019unsupervised}. 
For instance, Pumarola \emph{et al.} propose an unsupervised framework \cite{pumarola2018unsupervised} to generate person images, which induces some geometric errors as revealed in their paper.

Note that the aforementioned methods adopt human keypoints or skeleton as pose guidance, which are usually extracted by using OpenPose \cite{cao2017realtime}.
In addition, several works adopt DensePose \cite{neverova2018dense}, 3D pose \cite{li2019dense}, and segmented pose \cite{dong2018soft} to generate person images because they contain more information about body depth and part segmentation, producing better results with more texture details. 
However, the keypoint-based pose representation is much cheaper and more flexible than the DensePose, 3D pose, segmented pose representations, and can be more easily applied to practical applications.
Therefore, we favor keypoint-based pose representation in this paper.

\noindent \textbf{Image-Guidance Conditioning Schemes.}
Recently, there were proposed many schemes to incorporate the extra guidance (\emph{e.g.}, human poses \cite{ma2017pose,zhu2019progressive}, segmentation maps \cite{park2019semantic,tang2019multi,tang2020local}, facial landmarks \cite{tang2019cycle,zakharov2019few}, \emph{etc}) into an image-to-image translation model, which can be divided into four categories, 
\emph{i.e.}, input concatenation \cite{tang2019cycle,xian2018texturegan,zhang2017real}, feature concatenation \cite{ma2017pose,ma2018disentangled,esser2018variational,li2019dense,lai2018learning,li2019joint}, one-way guidance-to-image interaction \cite{siarohin2018deformable,park2019semantic,huang2017arbitrary,perez2018film}, two-way guidance-and-image interaction~\cite{zhu2019progressive,albahar2019guided,chi2019two}.

The most straightforward way of conditioning the guidance is to concatenate the input image and the guidance along the channel dimension.
For example, C2GAN \cite{tang2019cycle} takes the input person image and the targeted poses as input to output the corresponding targeted person images.
Instead of concatenating the guidance and the image at the input, several works \cite{ma2017pose,ma2018disentangled,esser2018variational} concatenate their feature representations at a certain layer.
For instance, PG2 \cite{ma2017pose} concatenates the embedded pose feature with the embedded image feature at the bottleneck fully connected layer. 
Another more general scheme is to use the guidance to guide the generation of the image.
For example, Siarohin \emph{et al.} \cite{siarohin2018deformable} first learn an affine transformation between the input and the target pose, then they use it to `move' the feature maps between the input image and the targeted image.
Unlike existing one-way guidance-to-image interaction schemes that allow information flow only from the guidance to the input image, a recent scheme, \emph{i.e.}, two-way guidance-and-image interaction, also considers the information flow from the input image back to the guidance \cite{zhu2019progressive,albahar2019guided}.
For example, Zhu \emph{et al.} \cite{zhu2019progressive} propose an attention-based GAN model to simultaneously update the person's appearance and shape features under the guidance of each other, and show that the proposed two-way guidance-and-image interaction strategy leads to better performance on person image generation tasks.

Contrary to the existing two-way guidance-and-image interaction schemes \cite{zhu2019progressive,albahar2019guided} that allow both the image and guidance to guide and update each other in a local way, we show that the proposed cross-conditioning strategy can further improve the performance of person image generation tasks.
\section{Xing Generative Adversarial Networks}
We start by presenting the details of the proposed XingGAN (Fig.~\ref{fig:method}) consisting of three parts, \emph{i.e.}, a Shape-guided Appearance-based generation (SA) branch modeling the person shape representation, an Appearance-guided Shape-based generation (AS) branch modeling the person appearance representation, and a Co-Attention Fusion (CAF) module for fusing these two branches. 
In the following, we first present the design of the two proposed generation branches, and then introduce the co-attention fusion module.
Lastly, we present the proposed two discriminators, the overall optimization objective and implementation details.

The inputs of the proposed Xing generator are the source image $I_s$, the source pose $P_s$, and the target pose $P_t$.
The goal is to translate the pose of the person in the source image $I_s$ from the source pose $P_s$ to the target pose $P_t$, thus synthesizing a photo-realistic person image $I_t^{'}$.
In this way, the source image $I_s$ provides the appearance information and the poses ($P_s$, $P_t$) provide the shape information to the Xing generator for synthesizing the desired person image. 

\begin{figure}[t]
	\centering
	\includegraphics[width=0.83\linewidth]{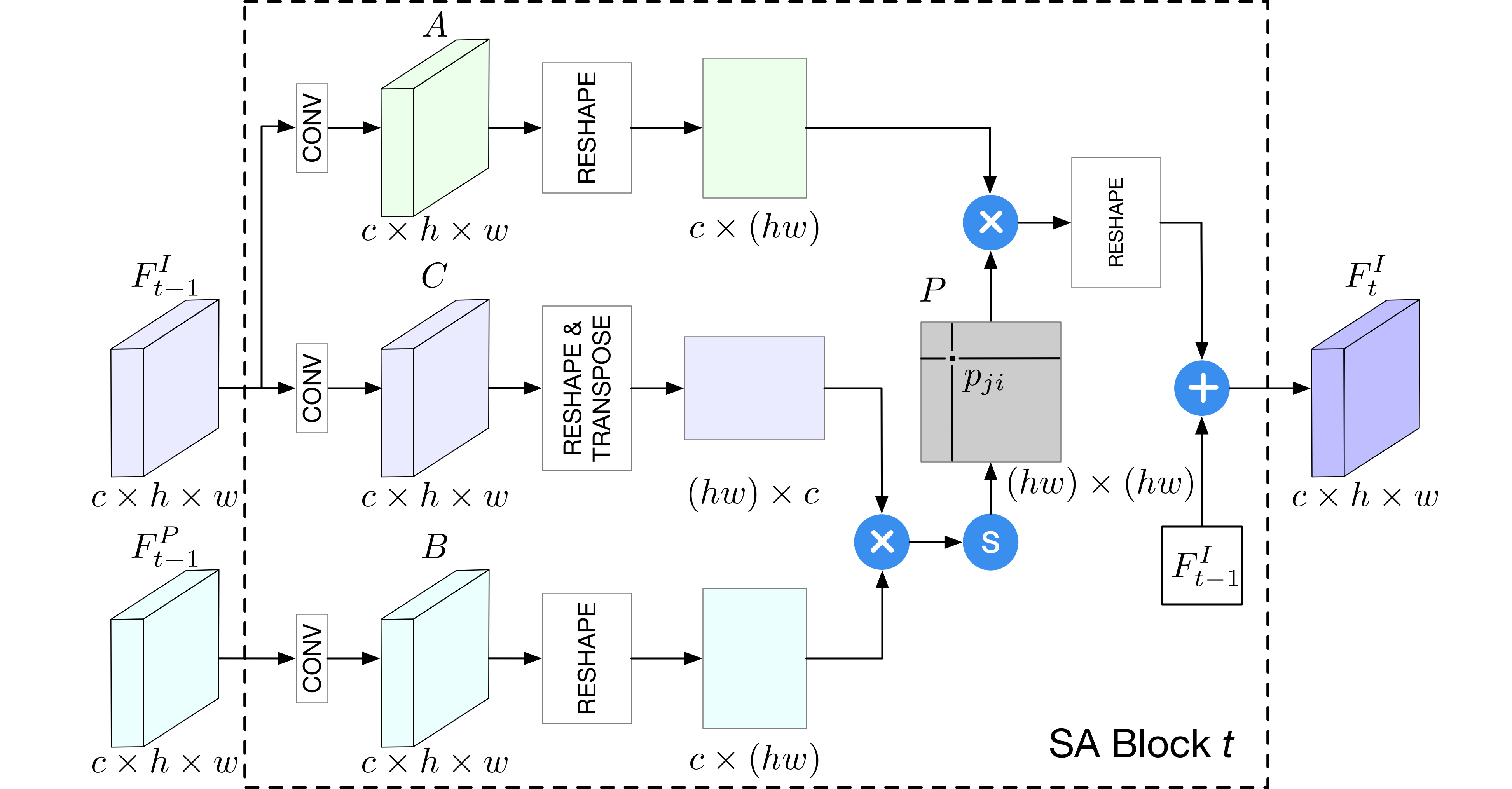}
	\caption{Structure of the proposed SA block which takes the previous appearance code $F_{t-1}^I$ and the previous shape code $F_{t-1}^P$ as input and obtains the appearance code $F_{t}^I$ in a crossed non-local way. The symbols $\oplus$, $\otimes$ and $\textcircled{s}$ and $\textcircled{c}$ denote element-wise addition, element-wise multiplication, Softmax activation, and channel-wise concatenation, respectively.}
	\label{fig:sa}
\end{figure}

\noindent \textbf{Shape-Guided Appearance-Based Generation.}
The proposed Shape-guided Appearance-based generation (SA) branch consists of an image encoder and a series of the proposed SA blocks.
The source image $I_s$ is first fed into the image encoder to produce the appearance code $F_0^I$, as shown in Fig. \ref{fig:method}.
The encoder consists of two convolutional layers in our experiments. 
The SA branch contains several cascaded SA blocks which progressively update the initial appearance code $F_0^I$ to the final appearance code $F_T^I$ under the guidance of the AS branch. 
As we can see in Fig.~\ref{fig:method}, all SA blocks have an identical network structure.
Consider the $t$-th block in Fig.~\ref{fig:sa}, whose inputs are the appearance code $F_{t-1}^I {\in} \mathbb{R}^{c \times h \times w}$ and the shape code $F_{t-1}^P {\in} \mathbb{R}^{c \times h \times w}$. The output is the refined appearance code $F_t^I {\in} \mathbb{R}^{c \times h \times w}$.
Specifically, given the appearance code $F_{t-1}^I$, we first feed it into a convolution layer to generate a new appearance code $C$, where $C {\in} \mathbb{R}^{c \times h \times w}$. 
Then we reshape $C$ to $\mathbb{R}^{c \times (hw)}$, where $n {=} hw$ is the number of pixels.
At the same time, the SA block receives the shape code $F_{t-1}^P$ from the AS branch, which is also fed into a convolution layer to produce a new shape code $B {\in} \mathbb{R}^{c \times h \times w}$ and then reshape to $\mathbb{R}^{c \times (hw)}$.
After that, we perform a matrix multiplication between the transpose of $C$ and $B$, and apply a Softmax layer to produce a correlation matrix $P {\in} \mathbb{R}^{(hw) \times (hw)}$,
\begin{equation}
\begin{aligned}
p_{ji} = \frac{{\rm exp} ( B_i  C_j)}{\sum_{i=1}^n {\rm exp}( B_i C_j)},
\label{eq:non_local1}
\end{aligned}
\end{equation}
where $p_{ji}$ measures the impact of the $i$-th position of $B$ on the $j$-th position of the appearance code $C$.
In this crossing way, the SA branch can capture more joint influence between the appearance code $F_{t-1}^I$ and shape code $F_{t-1}^P$, producing a richer appearance code~$F_{t}^I$.

Note that Eq.~\eqref{eq:non_local1} has a close relationship with the non-local operator proposed by Wang \emph{et al.} \cite{wang2018non}. 
The major difference is that the non-local operator in \cite{wang2018non} computes the pairwise similarity within the same feature map to incorporate global information, whereas the proposed crossing way computes the pairwise similarity between different feature maps, \emph{i.e.}, the person appearance and shape feature maps.

After that, we feed $F_{t-1}^I$ into a convolution layer to produce a new appearance code $A {\in} \mathbb{R}^{c \times h \times w}$ and reshape it to $\mathbb{R}^{c \times (h w)}$.
We then perform a matrix multiplication between $A$ and the transpose of $P$ and reshape the result to $\mathbb{R}^{c \times h \times w}$.
Finally, we multiply the result by a scale parameter $\alpha$ and conduct an element-wise sum operation with the original appearance code $F_{t-1}^I$ to obtain the refined appearance code $F_t^I \in \mathbb{R}^{c \times h \times w}$,
\begin{equation}
\begin{aligned}
F_t^I = \alpha \sum_{i=1}^{n}(p_{ji}  A_i) + F_{t-1}^I,
\end{aligned}
\end{equation}
where $\alpha$ is 0 in the beginning and but is gradually updated. By doing so, each position of the refined appearance code $F_t^I$ is a weighted sum of all positions of the shape code $F_{t-1}^P$ and the previous appearance code $F_{t-1}^I$.
Thus, it has a global contextual view between $F_{t-1}^P$ and $F_{t-1}^I$, and it selectively aggregates useful contexts according to the correlation matrix $P$. 

\noindent \textbf{Appearance-Guided Shape-Based Generation.}
In our preliminary experiments, we observe that only the SA generation branch is not sufficient to learn such a complex deformable translation process. Intuitively, since the shape features can guide the appearance features, we believe the appearance features can also be used to guide the shape features in turn.
Therefore, we also propose an Appearance-guided Shape-based generation (AS) branch.
The proposed AS branch mainly consists of a pose encoder and a sequence of AS blocks, as shown in Fig.~\ref{fig:method}.
The source pose $P_s$ and target pose $P_t$ are first concatenated along the channel dimension and then fed into the pose encoder to produce the initial shape representation $F_0^P$.
The pose encoder has the same network structure as the image encoder.
Note that to capture the dependency between the two poses, we only adopt one pose encoder.

\begin{figure}[t]
	\centering
	\includegraphics[width=0.78\linewidth]{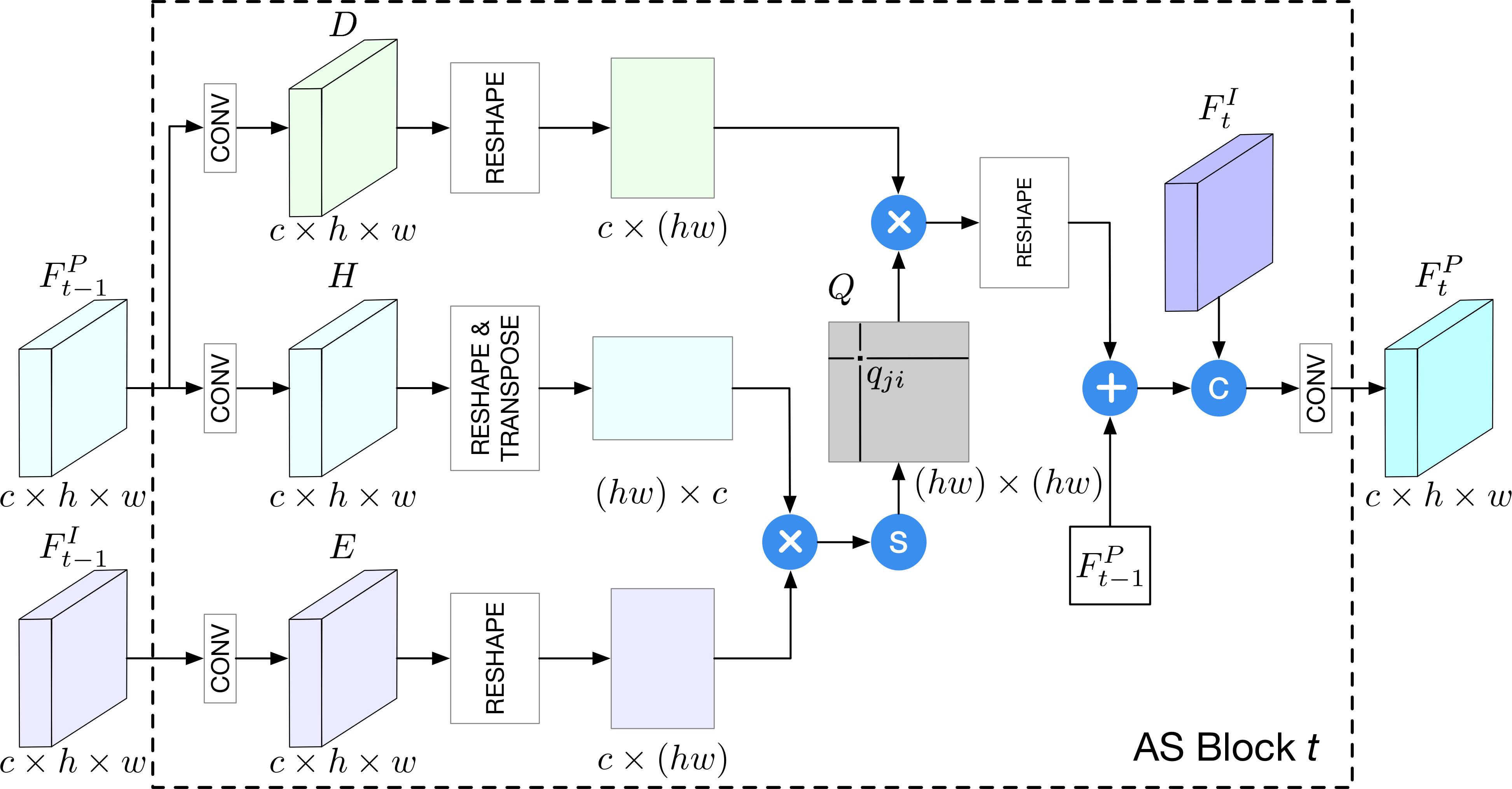}
	\caption{Structure of the proposed AS block, which takes the previous shape code $F_{t-1}^P$ and the previous appearance code $F_{t-1}^I$ as inputs and obtains the shape code $F_{t}^P$ in a crossing way. The symbols $\oplus$, $\otimes$ and $\textcircled{s}$ and $\textcircled{c}$ denote element-wise addition, element-wise multiplication, Softmax activation, and channel-wise concatenation, respectively.}
	\label{fig:as}
\end{figure}

The AS branch contains several cascaded AS blocks, which progressively update the initial shape code $F_0^P$ to the final shape code $F_T^P$ under the guidance of the SA branch. 
All AS blocks have the same network structure, as illustrated in Fig.~\ref{fig:method}.
Consider the $t$-th block in Fig. \ref{fig:as}, whose inputs are the shape code $F_{t-1}^P {\in} \mathbb{R}^{c \times h \times w}$ and the appearance code $F_{t-1}^I {\in} \mathbb{R}^{c \times h \times w}$. The output is the refined shape code $F_t^P {\in} \mathbb{R}^{c \times h \times w}$.

Specifically, given the shape code $F_{t-1}^P$, we first feed it into a convolution layer to generate a new shape code $H$, where $H {\in} \mathbb{R}^{c \times h \times w}$. 
We then reshape $H$ to $\mathbb{R}^{c \times (h w)}$.
At the same time, the AS block receives the appearance code $F_{t-1}^I$ from the SA branch, which is also fed into a convolution layer to produce a new appearance code $E$ and then reshape it to $\mathbb{R}^{c \times (h w)}$.
After that, we perform a matrix multiplication between the transpose of $H$ and $E$, and apply a Softmax layer to produce another correlation matrix $Q {\in} \mathbb{R}^{(h w) \times (h w)}$,
\begin{equation}
\begin{aligned}
q_{ji} = \frac{{\rm exp} (E_i  H_j)}{\sum_{i=1}^n {\rm exp}(E_i  H_j)},
\end{aligned}
\end{equation}
where $q_{ji}$ measures the impact of $i$-th position of $E$ on the $j$-th position of the shape code $H$. $n {=} h w$ is the number of pixels.

Meanwhile, we feed $F_{t-1}^P$ into a convolution layer to produce a new shape code $D {\in} \mathbb{R}^{c \times h \times w}$ and reshape it to $\mathbb{R}^{c \times (h  w)}$.
We then perform a matrix multiplication between $D$ and the transpose of $Q$ and reshape the result to $\mathbb{R}^{c \times h \times w}$.
Finally, we multiply the result by a scale parameter $\beta$ and conduct an element-wise sum operation with the original shape code $F_{t-1}^P$.
The result is then concatenated with the appearance code $F_t^I$ and fed into a convolution layer to obtain the updated shape code $F_t^P {\in} \mathbb{R}^{c \times h \times w}$,
\begin{equation}
\begin{aligned}
F_t^P = {\rm Concat}( \beta \sum_{i=1}^{n}(q_{ji} D_i) + F_{t-1}^P, F_t^I),
\end{aligned}
\end{equation}
where $\mathrm{Concat}(\cdot)$ denotes the channel-wise concatenation operation and $\beta$ is a parameter.
By doing so, each position in the refined shape code $F_t^P$ is a weighted sum of all positions in the appearance code $F_{t-1}^I$ and previous shape code $F_{t-1}^P$.

\noindent \textbf{Co-Attention Fusion.}
The proposed Co-Attention Fusion (CAF) module consists of two parts, \emph{i.e.}, generating intermediate results and co-attention maps.
These co-attention maps are used to spatially select from both the intermediate generations and the input image, and are combined to synthesize a final output.
This idea of the proposed CAF module comes from the multi-channel attention selection module in SelectionGAN \cite{tang2019multi}.
However, there are three differences: (i) We use two generation branches to generate intermediate results, \emph{i.e.}, SA branch and AS branch.
(ii) Attention maps are generated by the combination of both shape and appearance features, so the model learns more correlations between the two features.
(iii) We also produce the input attention map, which aims to select useful content from the input image for generating the final image.

We consider two directions to generate intermediate results.
One is generating multiple intermediate image synthesis results from the final appearance code $F_T^I$, and the other is generating multiple intermediate image synthesis results from the final shape code $F_T^P$.
Specifically, the appearance code $F_T^I$ is fed into a decoder to generate $N$ intermediate results $I^I{=}\{I_i^I\}_{i=1}^N$, and followed by a $\rm{Tanh}$ activation function.
Meanwhile, the final shape code $F_T^P$ is fed into another decoder to generate another $N$ intermediate results $I^P{=}\{I_i^P\}_{i=1}^{N}$, and also followed by a $\rm{Tanh}$ activation function.
Both can be formulated as,
\begin{equation}
\begin{aligned}
I_i^I = {}& {\rm Tanh}(F_T^I W_i^I +b_i^I),  & \quad   {\rm for}~i  =  1, \cdots, N \\
I_i^P ={} & {\rm Tanh}(F_T^P W_i^P +b_i^P),  & \quad   {\rm for}~i  =  1, \cdots, N
\end{aligned}
\end{equation}
where two convolution operations are performed with $N$ convolutional filters $\{W_i^I, b_i^I\}_{i=1}^{N}$ and $\{W_i^P, b_i^P\}_{i=1}^{N}$.
Thus, the $2N$ intermediate results and the input image $I_s$ can be regarded as the candidate image pool.

To generate the co-attention map which reflects the correlation between the appearance $F_T^I$ and shape $F_T^P$ codes, we first stack both $F_T^I$ and $F_T^P$ along the channel axes, and then feed them into a group of filters $\{W_i^A, b_i^A\}_{i=1}^{2N+1}$ to generate the corresponding $2N{+}1$ co-attention maps,
\begin{equation}
\begin{aligned}
I_i^A =  {\rm Softmax}({\rm Concat}(F_T^I, F_T^P) W_i^A +b_i^A),  & \quad   {\rm for}~i  =  1, \cdots, 2N{+}1 
\end{aligned}
\end{equation}
where ${\rm Softmax}$ is a channel-wise Softmax function used for the normalization, and $\mathrm{Concat}(\cdot)$ denotes the channel-wise concatenation operation.
Finally, the learned co-attention maps are used to perform a channel-wise selection from each intermediate generation and the input image as follows,
\begin{equation}
\begin{aligned}
I_t^{'} = (I_1^A \otimes I_1^I) \oplus \cdots (I_{2N}^A \otimes I_{2N}^P)  \oplus (I_{2N+1}^A \otimes I_s),
\end{aligned}
\end{equation}
where $I_t^{'}$ represents the final synthesized person image selected from the multiple diverse results and the input image. $\otimes$ and $\oplus$ denote the element-wise multiplication and addition, respectively. 

\noindent \textbf{Optimization Objective.}
We use three different losses as our full optimization objective, \emph{i.e.}, adversarial loss $\mathcal{L}_{gan}$, pixel loss $\mathcal{L}_{l1}$, and perceptual loss $\mathcal{L}_{p}$,
\begin{equation}
\begin{aligned}
\min_G \max_{D_I, D_P}  \mathcal{L} = \lambda_{gan} \mathcal{L}_{gan} + \lambda_{l1} \mathcal{L}_{l1} + \lambda_{p} \mathcal{L}_{p},
\label{eq:loss} 
\end{aligned}
\end{equation}
where $\lambda_{gan}$, $\lambda_{l1}$ and $\lambda_{p}$ are the weights, measuring corresponding contributions of each loss to the total loss $\mathcal{L}$.
The total adversarial loss is derived from the appearance-guided discriminator $D_I$ and the shape-guided discriminator $D_P$, which aims to judge how likely is that $I_t^{'}$ contains the same person in $I_s$ and how well $I_t^{'}$ aligns with the target pose $P_t$, respectively.
The $L1$ pixel loss is used to compute the difference between the generated image $I_t^{'}$ and the real target image $I_t$, \emph{i.e.}, $\mathcal{L}_{l1}{=} \vert\vert I_t-I_t^{'}\vert\vert_1$.
The perceptual loss $\mathcal{L}_{p}$ is used to reduce pose distortions and make the generated images look more natural and smooth, \emph{i.e.}, $\mathcal{L}_{p}{=}\vert\vert \phi(I_t)-\phi(I_t^{'})\vert\vert_1$, where $\phi$ denotes the outputs of several layers in the pre-trained VGG19 network \cite{simonyan2014very}.

\noindent \textbf{Implementation Details.}
We follow the training procedures of GANs and alternatively train the proposed Xing generator $G$ and two discriminators ($D_I$, $D_P$).
During training, $G$ takes $I_s$, $P_s$ and $P_t$ as input and outputs a translated person image $I_t^{'}$ with target pose $P_t$.
Specifically, $I_s$ is fed to the SA branch, and $P_s$, $P_t$ are fed to the AS branch.
For the adversarial training, ($I_s$, $I_t$) and ($I_s$, $I_t^{'}$) are fed to the appearance-guided discriminator $D_P$ for ensuring appearance consistency.
($P_t$, $I_t$) and ($P_t$, $I_t^{'}$) are fed to the shape-guided discriminator $D_P$ for ensuring shape consistency.

Adam optimizer \cite{kingma2014adam} is used to train the proposed XingGAN for around 90K iterations with $\beta_1{=}0.5$ and $\beta_2{=}0.999$.
We set $T{=}9$ in the proposed Xing generator and $N{=}10$ in the proposed co-attention fusion module on both datasets.
$\lambda_{gan}$, $\lambda_{l1}$ and $\lambda_{p}$ in Eq.~\eqref{eq:loss} are set to 5, 50 and 50, respectively.
For the decoders, the kernel size of convolutions for generating the intermediate images and co-attention maps are $3 {\times} 3$ and $1 {\times} 1$, respectively.

\section{Experiments}

\noindent \textbf{Datasets.}
We follow \cite{ma2017pose,siarohin2018deformable,zhu2019progressive} and conduct experiments on two challenging datasets, \emph{i.e.}, Market-1501 \cite{zheng2015scalable} and DeepFashion \cite{liu2016deepfashion}.
Images on Market-1501 and DeepFashion are rescaled to $128 {\times} 64$ and $256 {\times} 256$, respectively.
To generate human skeletons as training data, we employ OpenPose \cite{cao2017realtime} to extract human joints. 
In this way, both $P_s$ and $P_t$ consist of an 18-channel heat map encoding the positions of 18 joints of a human body.
We also filter out images where no human is detected.
Thus, we collect 101,966 training pairs and 8,570 testing pairs on DeepFashion.
For Market-1501, we have 263,632 training and 12,000 testing pairs. 
Note that to better evaluate the proposed XingGAN, the person identities of the training set do not overlap with those of the testing set.

\noindent \textbf{Evaluation Metrics.}
We follow \cite{ma2017pose,siarohin2018deformable,zhu2019progressive} and adopt Structure Similarity (SSIM) \cite{wang2004image}, Inception Score (IS) \cite{salimans2016improved}, and their masked versions, \emph{i.e.}, Mask-SSIM and Mask-IS, as the evaluation metrics.
Moreover, we adopt the PCKh score proposed in \cite{zhu2019progressive} to explicitly assess the shape consistency.

\begin{table*}[!t]
	\centering
	\caption{Quantitative results on Market-1501 and DeepFashion. For all metrics, higher is better. ($\ast$) denotes the results tested on our testing set.}
		\begin{tabular}{lcccccccc} \toprule
			\multirow{2}{*}{Method}  & \multicolumn{5}{c}{Market-1501} & \multicolumn{3}{c}{DeepFashion} \\ \cmidrule(lr){2-6} \cmidrule(lr){7-9} 
			 & SSIM & IS   & Mask-SSIM & Mask-IS  & PCKh  & SSIM  & IS  & PCKh      \\ \hline	
			PG2~\cite{ma2017pose}                                        & 0.253 & 3.460 & 0.792 & 3.435   & - & 0.762 & 3.090  & - \\
			DPIG~\cite{ma2018disentangled}                           & 0.099 & 3.483 & 0.614 & 3.491   & - & 0.614 & 3.228    & - \\
			PoseGAN~\cite{siarohin2018deformable}              & 0.290 & 3.185 & 0.805 & 3.502   & - & 0.756 & 3.439    & -\\ 
			C2GAN~\cite{tang2019cycle}                                & 0.282 & 3.349 & 0.811 & 3.510   & - & -     & -            & -\\ 
			BTF~\cite{albahar2019guided}                               & -     & -     & -     & -       & - & 0.767 & 3.220                   & -\\ \hline
			PG2$^\ast$~\cite{ma2017pose}                             & 0.261 & 3.495 & 0.782 & 3.367   &0.73 & 0.773 & 3.163  & 0.89 \\ 
			PoseGAN$^\ast$~\cite{siarohin2018deformable}    & 0.291 & 3.230 & 0.807 & 3.502   & \textbf{0.94} & 0.760 & 3.362 & 0.94 \\ 
			VUnet$^\ast$~\cite{esser2018variational}              & 0.266 & 2.965 & 0.793 & 3.549   & 0.92 & 0.763 & 3.440 & 0.93 \\
			PoseWarp$^\ast$~\cite{balakrishnan2018synthesizing} &  - & - & - & - & - & 0.764 & 3.368 & 0.93 \\
			CMA$^\ast$~\cite{chi2019two} &  - & - & - & - & - & 0.768 & 3.213 & 0.92 \\
			Pose-Transfer$^\ast$~\cite{zhu2019progressive}  & 0.311 & 3.323 & 0.811 & 3.773   & \textbf{0.94} & 0.773 & 3.209 & \textbf{0.96} \\  
			XingGAN (Ours)                                                    & \textbf{0.313} & \textbf{3.506} & \textbf{0.816} & \textbf{3.872} & 0.93    &  \textbf{0.778} & \textbf{3.476}  & 0.95 \\ \hline	
			Real Data                                                               & 1.000 & 3.890 & 1.000 & 3.706   & 1.00 & 1.000 & 4.053 & 1.00 \\	
			\bottomrule	
	\end{tabular}
	\label{tab:pose_reuslts}
\end{table*}

\begin{figure}[!t]
	\centering
	\includegraphics[width=1\linewidth]{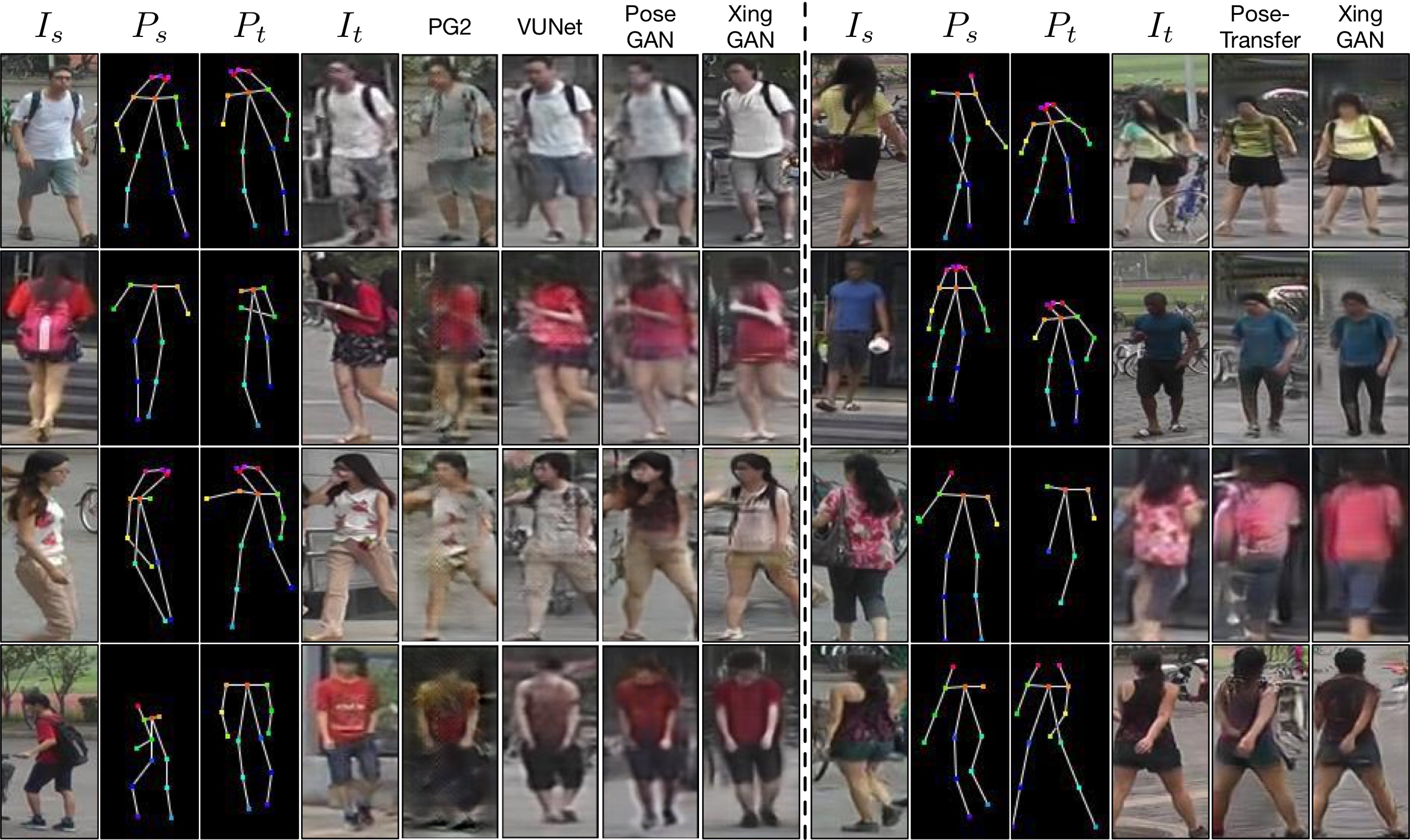}
	\caption{Qualitative comparison with PG2~\cite{ma2017pose}, VUnet~\cite{esser2018variational}, PoseGAN~\cite{siarohin2018deformable} and Pose-Transfer~\cite{zhu2019progressive} on Market-1501.}
	\label{fig:market_result}
\end{figure}

\begin{figure}[!t]
	\centering
	\includegraphics[width=1\linewidth]{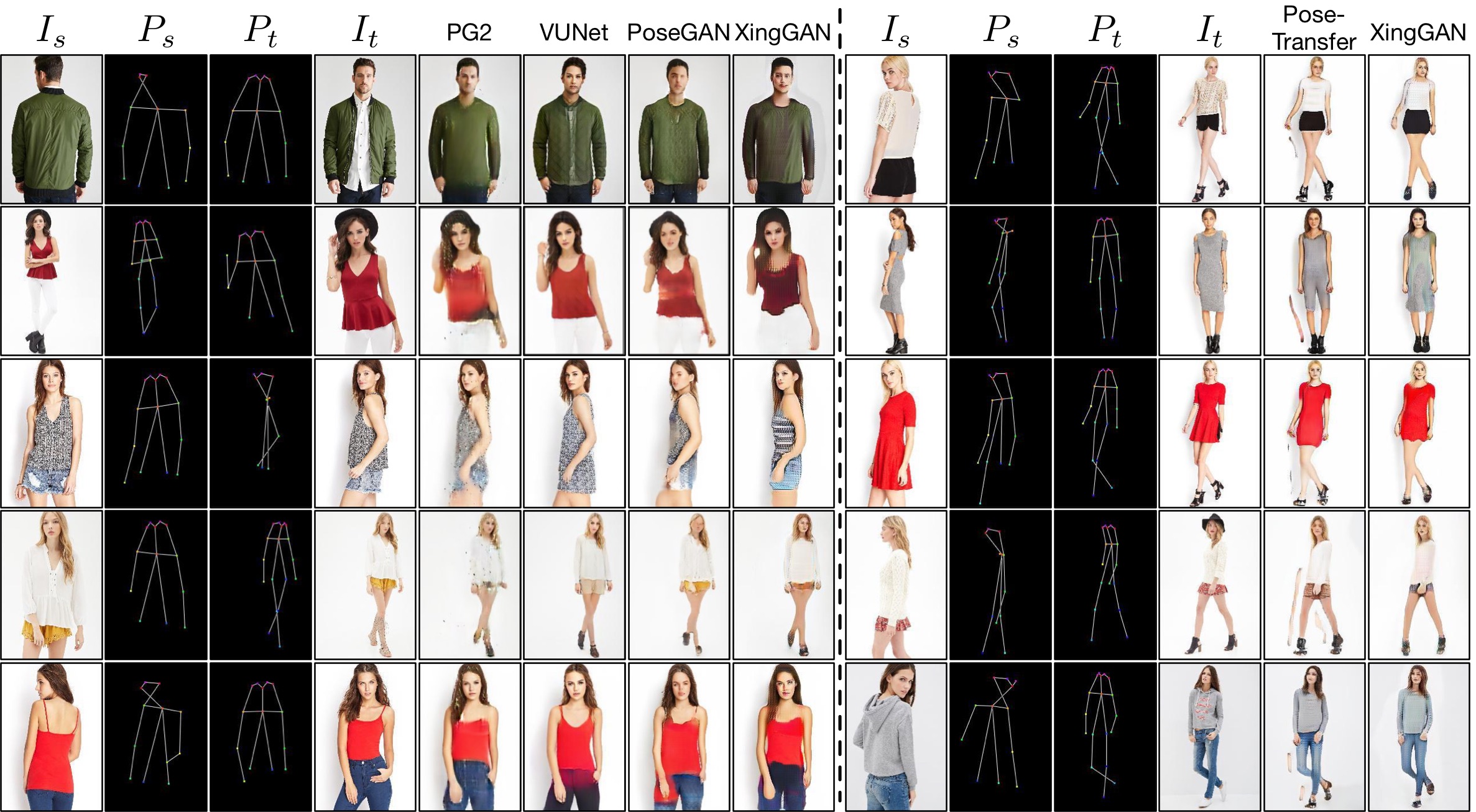}
	\caption{Qualitative comparison with PG2~\cite{ma2017pose}, VUnet~\cite{esser2018variational}, PoseGAN~\cite{siarohin2018deformable} and Pose-Transfer~\cite{zhu2019progressive} on DeepFashion.}
	\label{fig:fashion_result}
\end{figure}

\noindent \textbf{Quantitative Comparisons.}
We compare the proposed XingGAN with several leading methods, \emph{i.e.}, PG2~\cite{ma2017pose}, DPIG~\cite{ma2018disentangled}, VUnet~\cite{esser2018variational}, PoseGAN~\cite{siarohin2018deformable}, PoseWarp~\cite{balakrishnan2018synthesizing}, CMA~\cite{chi2019two}, C2GAN~\cite{tang2019cycle}, BTF~\cite{albahar2019guided} and Pose-Transfer~\cite{zhu2019progressive}.
Quantitative results measured by SSIM, IS, Mask-SSIM, Mask-IS, and PCKh metrics are shown in Table~\ref{tab:pose_reuslts}.
Note that previous works \cite{ma2017pose,siarohin2018deformable} did not release the train/test split, thus we use their well-trained models and re-evaluate their performance on our testing set as in Pose-Transfer~\cite{zhu2019progressive}.
Although our testing set inevitably includes some of their training samples, XingGAN still achieves the best results in terms of SSIM, IS, Mask-SSIM, and Mask-IS metrics on both datasets.
For the PCKh metric, \cite{zhu2019progressive} obtains slightly better results than XingGAN.
However, we observe that the images generated by XingGAN are more realistic and have less visual artifacts than those generated by \cite{zhu2019progressive}  (see Fig.~\ref{fig:market_result} and~\ref{fig:fashion_result}).

\noindent \textbf{Qualitative Comparisons.}
Results compared with PG2~\cite{ma2017pose}, VUnet~\cite{esser2018variational} and PoseGAN~\cite{siarohin2018deformable} are shown on the left of Fig. \ref{fig:market_result} and~\ref{fig:fashion_result}. 
We can see that the proposed XingGAN achieves much better results than PG2, VUnet, and PoseGAN on both datasets, especially at appearance details and the integrity of generated persons.
Moreover, to evaluate the effectiveness of XingGAN, we compare it with a stronger baseline, \emph{i.e.}, Pose-Transfer \cite{zhu2019progressive}.
Results are shown on the right of Fig.~\ref{fig:market_result} and~\ref{fig:fashion_result}.
We can see that XingGAN also generates much better person images having fewer visual artifacts than Pose-Transfer.
For instance, Pose-Transfer \cite{zhu2019progressive} always generates a lot of visual artifacts in the background as shown in Fig.~\ref{fig:fashion_result}.

\noindent \textbf{Human Evaluation.}
We follow the evaluation protocol of \cite{ma2017pose,siarohin2018deformable,zhu2019progressive} and recruited 30 volunteers to conduct a user study.
Participants were shown a sequence of images and asked to give an instant judgment about each image within a second.
Specifically, we randomly select 55 real and 55 fake images (generated by our model) and shuffle them.
The first 10 of them are used for practice and the remaining 100 images are used for evaluation.
Results compared with PG2~\cite{ma2017pose}, PoseGAN~\cite{siarohin2018deformable}, Pose-Transfer~\cite{zhu2019progressive} and C2GAN \cite{tang2019cycle} are shown in Table \ref{tab:pose_ruser}.
We observe that the proposed XingGAN achieves the best results on all measurements compared with the leading methods, further validating that the generated images by our model are more sharp and photo-realistic.

\begin{table}[!t]
	\centering
	\caption{User study of person image generation (\%). R2G means the percentage of real images rated as generated w.r.t. all real images. G2R means the percentage of generated images rated as real w.r.t. all generated images. The results of other methods are reported from their papers.}
		\begin{tabular}{lccccccc} \toprule
			\multirow{2}{*}{Method}  & \multicolumn{2}{c}{Market-1501} & \multicolumn{2}{c}{DeepFashion} \\ \cmidrule(lr){2-3} \cmidrule(lr){4-5} 
			& R2G & G2R & R2G & G2R    \\ \hline	
			PG2~\cite{ma2017pose}                              & 11.2  & 5.5    & 9.2   & 14.9 \\
			PoseGAN~\cite{siarohin2018deformable}    & 22.67 & 50.24  & 12.42 & 24.61 \\ 
			C2GAN~\cite{tang2019cycle}                      & 23.20 & 46.70  & -     & -     \\
			Pose-Transfer~\cite{zhu2019progressive}   & 32.23 & 63.47  & 19.14 & 31.78 \\  
			XingGAN (Ours)                                         & \textbf{35.28} & \textbf{65.16} & \textbf{21.61} & \textbf{33.75} \\	
			\bottomrule	
	\end{tabular}
	\label{tab:pose_ruser}
\end{table}

\noindent \textbf{Variants of XingGAN.}
We conduct extensive ablation studies on Market-1501~\cite{zheng2015scalable} to evaluate different components of our XingGAN.
XingGAN has four baselines as shown in Table~\ref{tab:abla}: (i) `SA' means only using the proposed Shape-guided Appearance-based generation branch. (ii) `AS' means only adopting the proposed Appearance-guided Shape-based generation branch. (iii) `SA+AS' combines both branches to produce the final person images. (iv) `SA+AS+CAF' is our full model and employs the proposed Co-Attention Fusion module.

\noindent \textbf{Effect of Dual-Branch Generation.}
The results of the ablation study are shown in Table \ref{tab:abla}. 
We see that the proposed SA branch achieves only 0.239 and 0.768 in SSIM and Mask-SSIM, respectively.
When we only use the proposed AS branch, the values of SSIM and Mask-SSIM are improved to 0.286 and 0.798, respectively.
Thus we conclude that the AS branch is more effective than the SA branch for generating photo-realistic person images.
The AS branch takes the person poses as input and aims to learn person appearance representations, while the SA branch takes the person image as input and targets to learn person shape representations.
Learning the appearance representations is much easier than learning the shape representations since there are shape deformations between the input person image and the desired person image, leading the AS branch to achieve better results than the SA branch.

\begin{table}[!t]
	\centering
	\caption{Quantitative comparison of different variants of the proposed XingGAN on Market-1501. For all metrics, higher is better. `SA', `AS' and `CAF' stand for the proposed SA branch, AS branch and co-attention fusion module, respectively.}
		\begin{tabular}{lcccc} \toprule
			Variants of XingGAN & IS & Mask-IS & SSIM & Mask-SSIM    \\ \hline	
			SA & \textbf{3.849} & 3.645 & 0.239 & 0.768 \\ 
			AS & 3.796 & 3.810 & 0.286 & 0.798 \\
			SA + AS & 3.558 & 3.807 & 0.310 & 0.807 \\
			SA + AS + CAF (Full)       & 3.506 & \textbf{3.872} & \textbf{0.313} & \textbf{0.816} \\
			\bottomrule	
	\end{tabular}
	\label{tab:abla}
\end{table}

\begin{figure}[!t]
	\centering
	\includegraphics[width=1\linewidth]{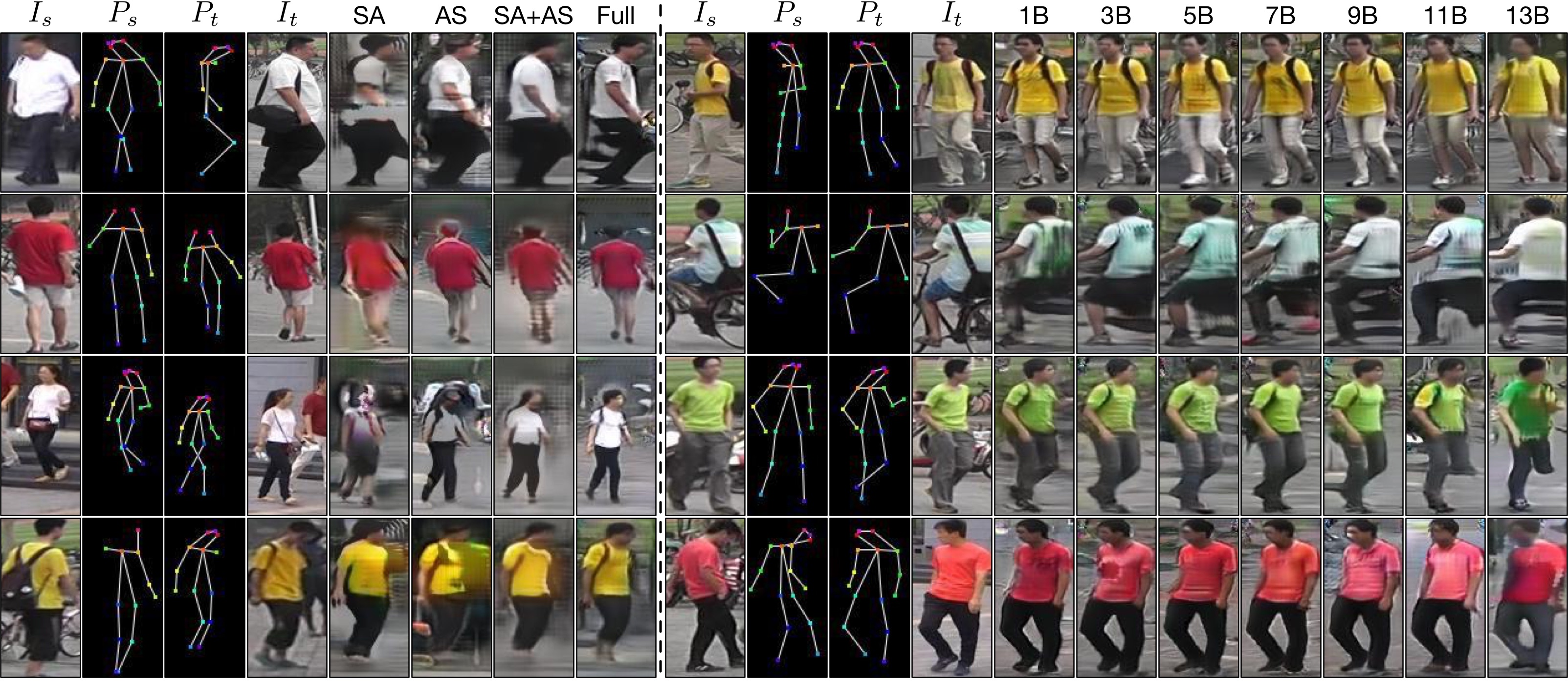}
	\caption{Ablation study of the proposed XingGAN on Market-1501. (\textit{left}) Results of different variants of the proposed XingGAN. (\textit{right}) Results of varying the number of the proposed Xing blocks. `B' stands for the proposed Xing Blocks.}
	\label{fig:ablation}
\end{figure}

Next, when adopting the combination of the proposed SA and AS branches, the performance in terms of SSIM and Mask-SSIM further boosts.
However, the results in terms of IS and Mask-IS do not decline too much.
Moreover, Fig.~\ref{fig:ablation} (\textit{left}) shows some qualitative examples of the ablation study.
We observe that the visualization results of `SA', `AS', and `SA+AS' are consistent with the quantitative results.
Therefore, both quantitative and qualitative results confirm the effectiveness of the proposed dual-branch generation strategy.

\noindent \textbf{Effect of Co-Attention Fusion.}
`SA+AS+CAF' outperforms the `SA+AS' baseline with around 0.065, 0.003, and 0.009 gain on Mask-IS, SSIM, and Mask-SSIM, respectively. 
This means that the proposed co-attention fusion model indeed learns more correlations between the appearance and shape representations for generating the targeted person images, confirming our design motivation.
Moreover, the proposed CAF module obviously improves the quality of the visualization results, as shown in the column `Full' of  Fig.~\ref{fig:ablation}.

\begin{figure}[!t]
	\centering
	\includegraphics[width=1\linewidth]{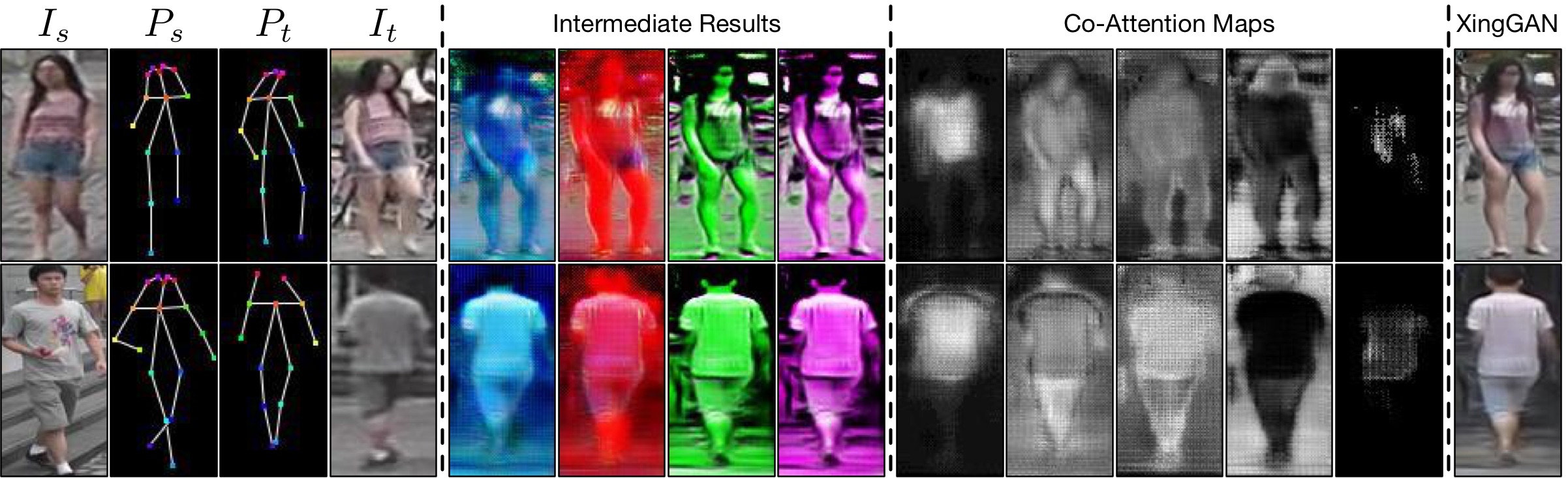}
	\caption{Visualization of intermediate results and co-attention maps generated by the proposed XingGAN on Market-1501. We randomly show four intermediate results, the corresponding four co-attention maps and the input attention map. Attention maps are normalized for better visualization.}
	\label{fig:attention_map}
\end{figure}

Lastly, we show the learned co-attention maps and the generated intermediate results.
These co-attention maps are complementary, which could be qualitatively verified by visualizing the results in Fig.~\ref{fig:attention_map}. 
It is clear that they have learned different activated content between the generated intermediate results and the input image for generating the final person images.

\begin{table}[!t]
	\centering
	\caption{Quantitative comparison and ablation study of the proposed Xing generator on Market-1501. For all metrics, higher is better. }
		\begin{tabular}{lcccc} \toprule
			Method                      & IS    & Mask-IS & SSIM   & Mask-SSIM    \\ \hline
			Xing Generator (1 blocks) & 3.378 & 3.713   & 0.310  & 0.812 \\
			Xing Generator (3 blocks)  & 3.241 & 3.866   & \textbf{0.316}  & 0.813 \\
			Xing Generator (5 blocks)   & 3.292 & 3.860     & 0.313  & 0.812 \\ 
			Xing Generator (7 blocks)   & 3.293 & 3.871     & 0.310  & 0.810 \\
			Xing Generator (9 blocks) & 3.506 & \textbf{3.872}     & 0.313  & \textbf{0.816}  \\
			Xing Generator (11 blocks)  & 3.428 & 3.712     & 0.286  & 0.793 \\
			Xing Generator (13 blocks)  & \textbf{3.708}    & 3.679     & 0.257  & 0.774 \\ \hline
			Resnet Generator (5 blocks) & 3.236 & 3.807     & 0.297  & 0.802 \\
			Resnet Generator (9 blocks) & 3.077 & 3.862     & 0.301 &  0.802 \\
			Resnet Generator (13 blocks)& 3.134 &  3.731    & 0.300 & 0.797 \\ \hline
			PATN Generator (5 blocks)   & 3.273 & 3.870     & 0.309 &  0.809 \\
			PATN Generator (9 blocks)   &  3.323 & 3.773    & 0.311 & 0.811 \\
			PATN Generator (13 blocks)  & 3.274 & 3.797     & 0.314 & 0.808 \\ 
			\bottomrule	
	\end{tabular}
	\label{tab:ablation_blocks}
\end{table}

\noindent \textbf{Effect of The Xing Generator.}
The proposed Xing generator has two important network designs. 
One is the carefully designed Xing block, consisting of two sub-blocks, \emph{i.e.}, SA block and AS block. 
The Xing blocks jointly model both shape and appearance representations in a crossing way and enjoying the mutually improved benefits from each other.
The other one is the cascaded network design, which deals with the complex and deformable translation problem progressively.
Thus, we further conduct two experiments, one is to show the advantage of the progressive generation strategy by varying the number of the proposed Xing blocks, and the other is to explore the advantage of the Xing block by replacing it with the residual block \cite{johnson2016perceptual} and PATB \cite{zhu2019progressive} resulting in two generators named Resnet generator and PATN generator in Table \ref{tab:ablation_blocks}, respectively.

Quantitative and qualitative results are shown in Table \ref{tab:ablation_blocks} and Fig. \ref{fig:ablation} (\textit{right}).
We observe that the proposed Xing generator with 9 blocks works the best. 
However, increasing the number of blocks further reduces generation performance.
This could be attributed to the proposed Xing block. Only a few blocks are needed to capture the useful appearance and shape representations and the connection between them.
Thus, we adopt 9 Xing blocks as default in our experiments for both datasets.
Moreover, we see that the proposed Xing generator with only 5 Xing blocks outperforms both ResNet and PATN generators with 13 blocks on most metrics, which further certifies that our Xing generator has a good appearance and shape modeling capabilities with a very few blocks.

\section{Conclusions}
We propose a novel XingGAN for the challenging person image generation task.
It uses cascaded guidance with two different generation branches, and learns a deformable translation mapping from both person shape and appearance features.
Moreover, we propose two novel blocks to effectively update person shape and appearance features in a crossing way.
Extensive experiments based on human judgments and automatic evaluation metrics show that XingGAN achieves new state-of-the-art results on two challenging datasets.
Lastly, we believe that the proposed blocks and the XingGAN framework can be easily extended to address other GAN-based generation and even multi-modality fusion tasks.

\noindent \textbf{Acknowledgment:} This work has been partially supported by the Italy-China collaboration project TALENT.

\clearpage
%
%
\bibliographystyle{splncs04}
\bibliography{5192}

\clearpage
This document  provides additional experimental results on the person image generation task. 
First, we compare the proposed XingGAN with  the most state-of-the-art method Pose-Transfer \cite{zhu2019progressive} (Sec.~\ref{sec:1}).
Additionally, we show more  ablation results of the proposed XingGAN (Sec.~\ref{sec:2}). 
Lastly, we also provide the visualization results of the generated co-attention maps (Sec.~\ref{sec:3}). 

\section{State-of-the-Art Comparisons}
\label{sec:1}
\begin{figure*}[t] 
	\centering
	\includegraphics[width=1\linewidth]{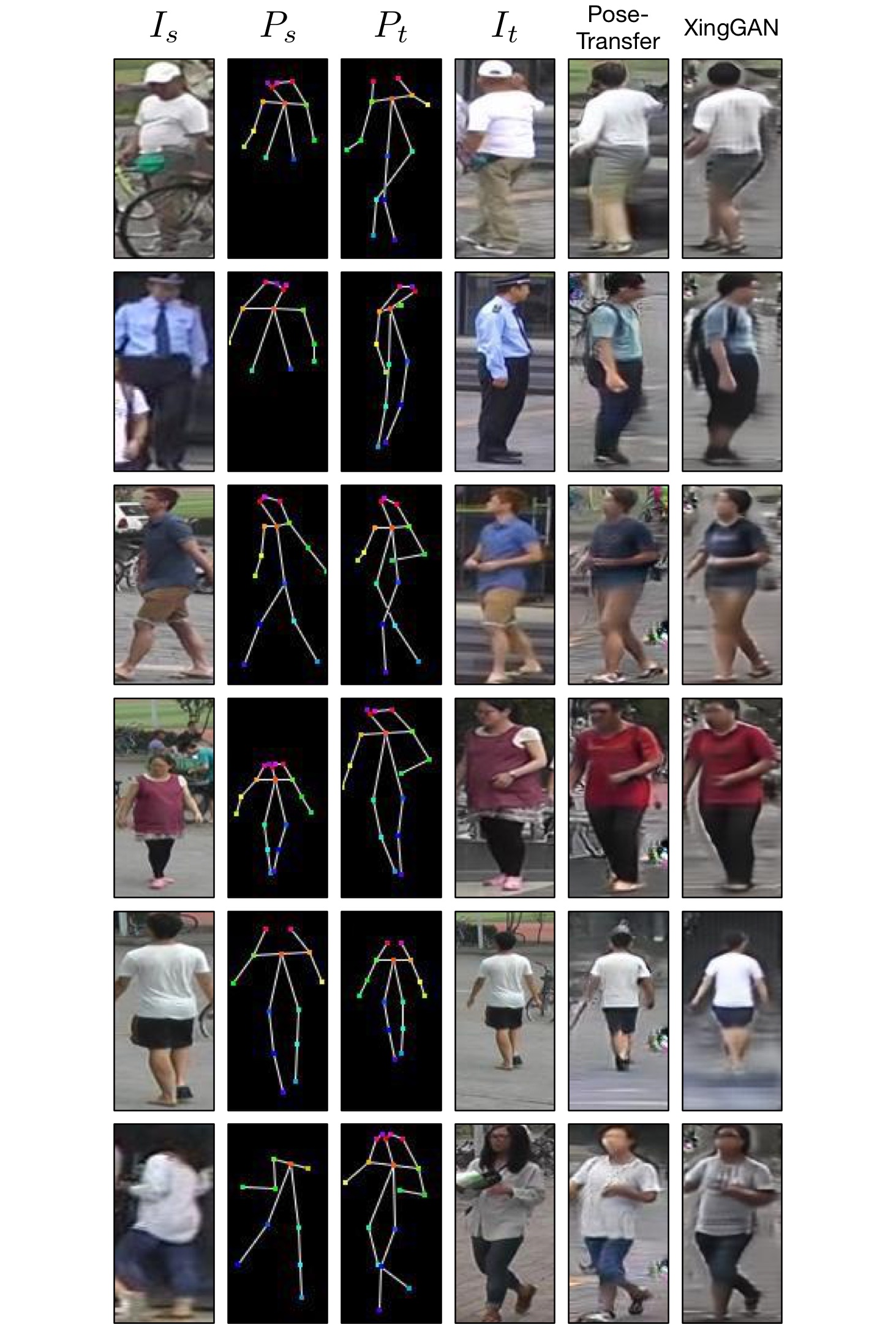}
	\caption{Qualitative comparison with Pose-Transfer~\cite{zhu2019progressive} on Market-1501.
	}
	\label{fig:supp_market}
\end{figure*}

\begin{figure*}[t] 
	\centering
	\includegraphics[width=1\linewidth]{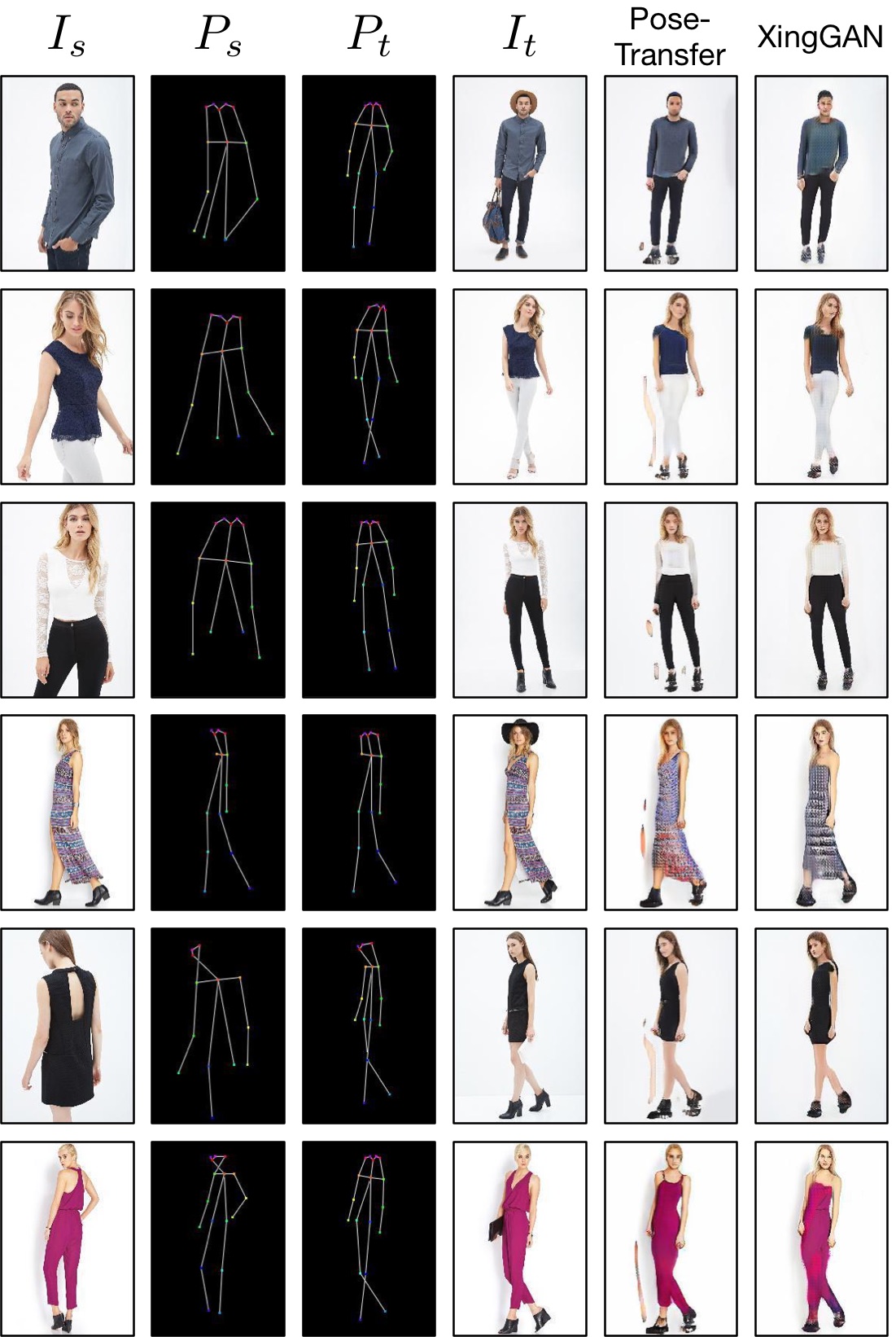}
	\caption{Qualitative comparison with Pose-Transfer~\cite{zhu2019progressive} on DeepFashion.
	}
	\label{fig:supp_fashion}
\end{figure*}

In Fig.~\ref{fig:supp_market} and \ref{fig:supp_fashion}, we provide more generation results of the proposed XingGAN  and Pose-Transfer \cite{zhu2019progressive} on both  the Market-1501 \cite{zheng2015scalable} and DeepFashion \cite{liu2016deepfashion} datasets. Note that  we generated the results of Pose-Transfer \cite{zhu2019progressive} using the well-trained models provided by the authors\footnote{\url{https://github.com/tengteng95/Pose-Transfer}} for fair comparisons. We observe that the proposed XingGAN consistently achieves  photo-realistic results with fewer visual artifacts than Pose-Transfer on both challenging datasets. 

\section{More Ablation Results}
\label{sec:2}
\begin{figure*}[t] 
	\centering
	\includegraphics[width=1\linewidth]{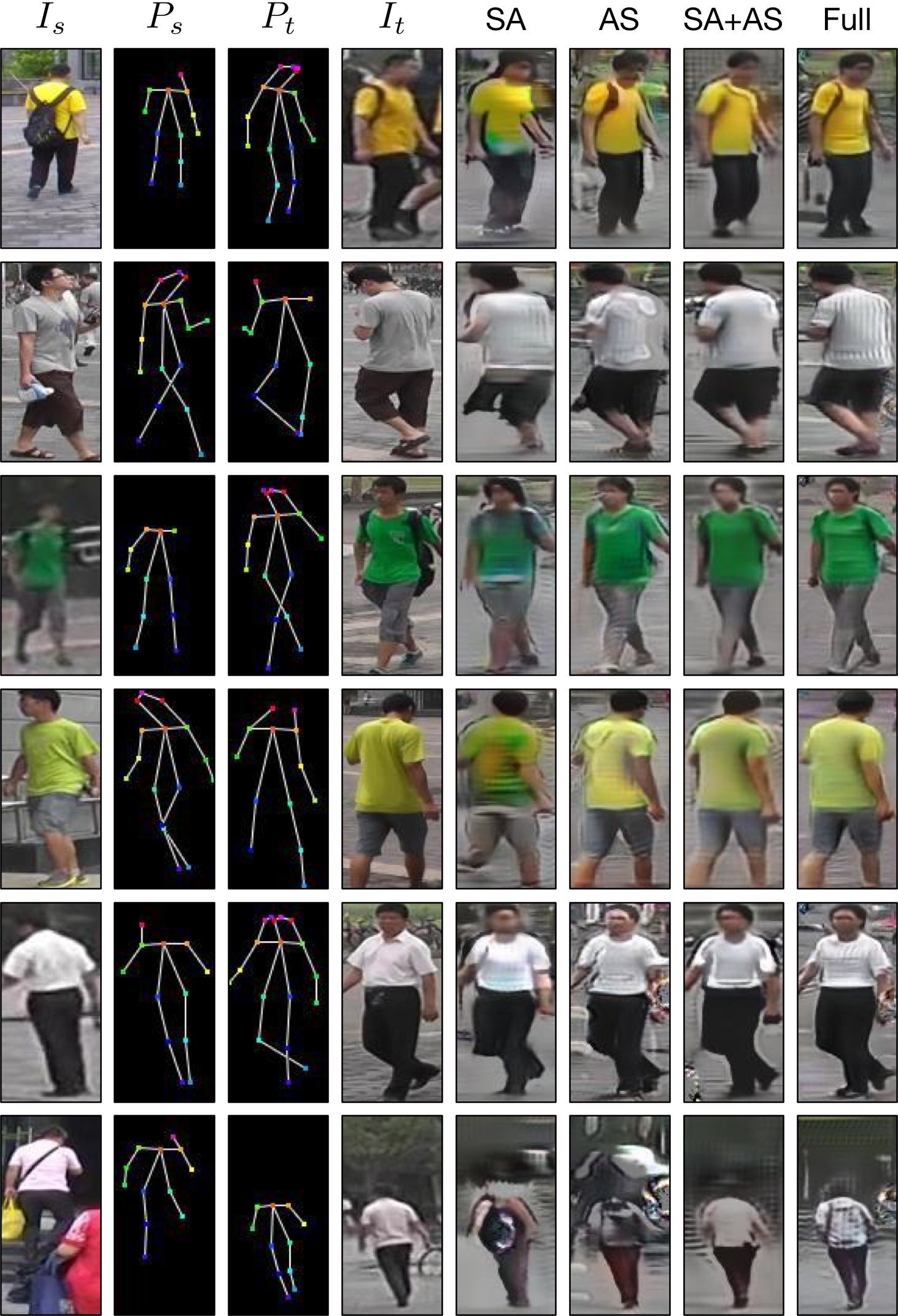}\\ 
	\caption{Ablation study results of different variants of the proposed XingGAN on Market-1501.
	}
	\label{fig:supp_ablation1}
\end{figure*}

In Fig.~\ref{fig:supp_ablation1}, we provide more qualitative ablation comparisons of the proposed XingGAN on Market-1501. 
These results further demonstrate the advantage of each component of the proposed XingGAN.
Moreover, we observe that our full model consistently generates more coherent and natural person images.

\begin{figure*}[t] 
	\centering
	\includegraphics[width=1\linewidth]{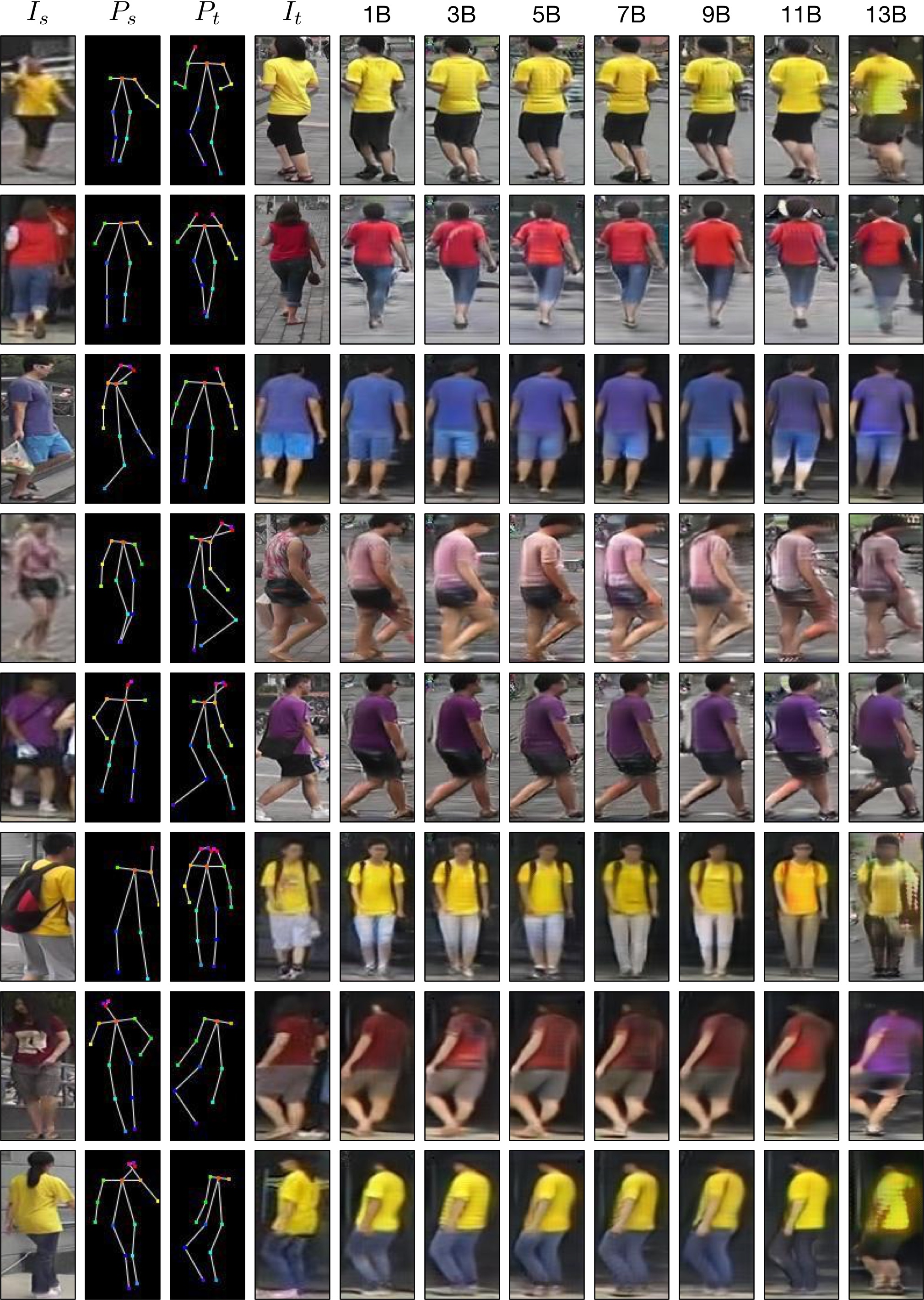}\\ 
	\caption{Ablation study results of varying the number of the proposed Xing blocks on Market-1501.  `B' stands for the proposed Xing Blocks.
	}
	\label{fig:supp_ablation2}
\end{figure*}

In Fig. \ref{fig:supp_ablation2}, we show the results of varying the number of the proposed Xing blocks on Market-1501.
We observe that adopting  about 9 Xing blocks makes the generated person images more natural and realistic, revealing the benefits of our progressive generation strategy.

\section{Visualization of Co-Attention Maps}
\label{sec:3}
\begin{figure*}[t] 
	\centering
	\includegraphics[width=1\linewidth]{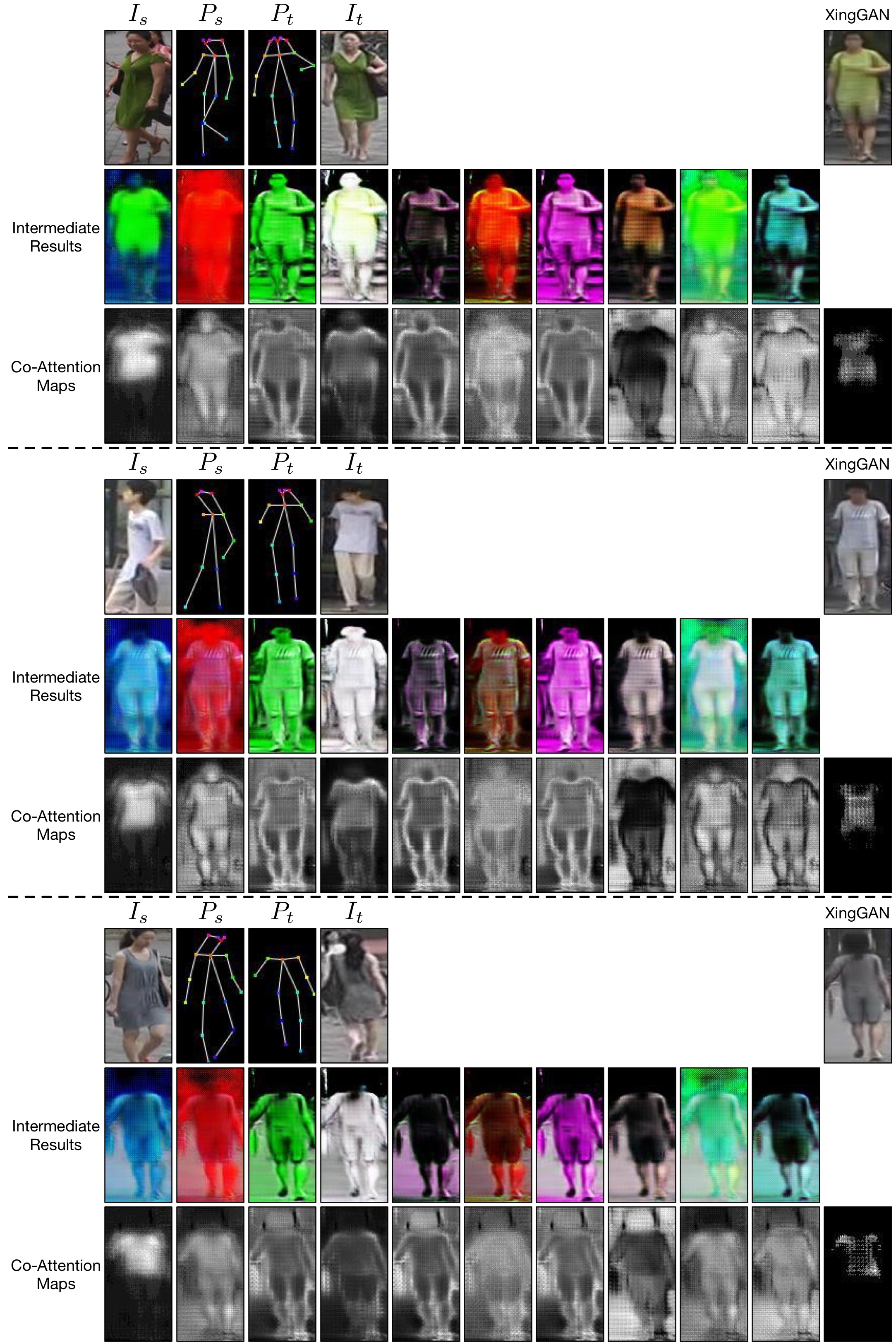}
	\caption{Visualization of intermediate results and co-attention maps generated by the proposed XingGAN on Market-1501. 
	}
	\label{fig:supp_attention}
\end{figure*}

We also provide more visualization results of the generated co-attention maps and intermediate results in Fig. \ref{fig:supp_attention}.
We show 10 randomly chosen intermediate results, their corresponding 10 co-attention maps, and the input attention map.
It is clear that these co-attention maps have learned different activated content between the generated intermediate results and the input image for generating the final person images, revealing the effectiveness of the proposed co-attention fusion  module.

\end{document}